\documentclass{esannV2}
\usepackage[dvips]{graphicx}
\usepackage[latin1]{inputenc}
\usepackage{multirow}
\usepackage{amssymb,amsmath,array}
\usepackage{booktabs}

\newcommand{\bp}{\mathbf{p}}

\newcommand{\bu}{\mathbf{u}}
\newcommand{\bw}{\mathbf{w}}

\newcommand{\balpha}{\boldsymbol{\alpha}}

\newcommand{\Ltrain}{\mathcal{L}_{train}}
\newcommand{\Lval}{\mathcal{L}_{val}}

\begin{document}
%style file for ESANN manuscripts
\title{A Lightweight Neural Architecture Search Model for Medical Image Classification}

%***********************************************************************
% AUTHORS INFORMATION AREA
%***********************************************************************
\author{Lunchen Xie$^{1, 2}$, Eugenio Lomurno$^2$, Matteo Gambella$^2$, Danilo Ardagna$^2$, \\Manuel Roveri$^2$, Matteo Matteucci$^2$, and Qingjiang Shi$^1$
\thanks{This paper is supported by the FAIR project, funded by the NextGenerationEU program within the PNRR-PE-AI scheme (M4C2, investment 1.3, line on Artificial Intelligence).
This paper is also supported by China Scholarship Council (CSC).}
%
% Optional short acknowledgment: remove next line if non-needed
% \thanks{This is an optional funding source acknowledgement.}
%
% DO NOT MODIFY THE FOLLOWING '\vspace' ARGUMENT
\vspace{.3cm}\\
$^1$Tongji University, Shanghai, China, \\School of Software Engineering
\vspace{.1cm}\\
$^2$Politecnico di Milano, Milan, Italy, \\Department of Electronics, Information, and Bioengineering
}
%***********************************************************************
% END OF AUTHORS INFORMATION AREA
%***********************************************************************

\maketitle
\vspace{-7pt}
\begin{abstract}
Accurate classification of medical images is essential for modern diagnostics. Deep learning advancements led clinicians to increasingly use sophisticated models to make faster and more accurate decisions, sometimes replacing human judgment. However, model development is costly and repetitive. Neural Architecture Search (NAS) provides solutions by automating the design of deep learning architectures. This paper presents ZO-DARTS+, a differentiable NAS algorithm that improves search efficiency through a novel method of generating sparse probabilities by bi-level optimization. Experiments on five public medical datasets show that ZO-DARTS+ matches the accuracy of state-of-the-art solutions while reducing search times by up to three times.
\end{abstract}
\vspace{-7pt}

\section{Introduction}\label{sec1}
\vspace{-5pt}

Accompanying the advancement of deep learning models, their application in medical image processing has become critically important in contemporary diagnostic processes. In fact, this technology provides experts with the ability to detect diseases earlier and with greater accuracy than previously possible. Although this revolution spans multiple data types and tasks, the design of optimal models remains costly. While image processing models are well established, achieving peak performance requires careful selection of appropriate structures and operations. In addition, the significant heterogeneity and variable quality of medical images necessitate domain-specific adaptation of these models, which requires significant human effort and expertise in neural network design.

To alleviate the burden of repetitive and demanding tasks from humans, the Neural Architecture Search (NAS) paradigm has been proposed and is nowadays considered a silver bullet for automating the model design process~\cite{nasframework}. As this approach is relatively new to the field of medical diagnosis, it holds considerable potential for application in medical image processing, particularly in classification tasks. Moreover, given the complexity of the initial NAS techniques, novel trends concerning efficient search strategies and hardware-aware techniques are currently being developed to enhance the performance and accessibility of deep learning solutions~\cite{falanti2023popnasv3, gambella_cnas_2022, kang2023neural}.

The main contribution of this paper is the introduction of ZO-DARTS+, an efficient and accurate differentiable NAS algorithm tailored for medical image classification. This new algorithm extends ZO-DARTS~\cite{xie2023zo} by incorporating sparsemax~\cite{martins2016softmax} along with an appropriate annealing strategy. These enhancements allow the algorithm to generate operation probabilities with sparse values during the search, improving interpretability. In addition, ZO-DARTS+ converges faster than its predecessor, reducing the average search time by 17.2\% without sacrificing, and often improving, the accuracy of the resulting architectures. When compared with other state-of-the-art solutions, ZO-DARTS+ achieves comparable levels of accuracy, while reducing the search time by up to three times.
\vspace{-10pt}
\section{Background}\label{sec2}
\vspace{-5pt}

The cell-based approach is prevalent in NAS to reduce search space as it can shrink the search space from model-level to cell-level. Inside each cell, operations between every $i^{th}$ and $j^{th}$ data block utilise a shared candidate set $\mathcal{O}=\{o^{(i,j)}\}$. DARTS introduces architecture parameters $\balpha=\{\balpha^{(i,j)}\}$, which are continuous and form a weighted sum of the operations~\cite{liu2018darts}. Each $\balpha^{(i,j)}\in\mathbb{R}^{|\mathcal{O}|}$ acts as the weight vector for the operation set $\{o^{(i,j)}\}$ and has to be optimized. The mixed operation is defined by $\bar{o}^{(i,j)}(x)=\sum_{o\in\mathcal{O}}\frac{\exp(\balpha_o^{(i,j)})}{\sum_{o'\in\mathcal{O}}\exp(\balpha_{o'}^{(i,j)})}o(x)$, normalized by the softmax function. This introduces a bi-level optimization problem with $\boldsymbol{\alpha}$ as the upper-level variable and model parameters $\bw$ as the lower-level variable:
\vspace{-3pt}
\begin{equation}\label{eq1}
\begin{aligned}
&\min_{\balpha}\quad F(\balpha)=\Lval(\mathbf{w}^*(\balpha),\balpha)\
&s.t.\quad\bw^*(\balpha)=\arg\min_{\bw}\Ltrain(\bw,\balpha).
\end{aligned}
\end{equation}
Here, $\Lval(\bw^*,\balpha)$ and $\Ltrain(\mathbf{w},\balpha)$ are loss functions on the validation and training datasets respectively. The final discrete architecture is derived by applying argmax over $\{\balpha^{(i,j)}\}$.

To solve Eq.\eqref{eq1} accurately, the analytical gradient of $F(\balpha)$ is derived using the implicit function theorem~\cite{lorraine2020optimizing}:
\vspace{-3pt}
\begin{equation}\label{eq2}
\nabla_{\!\balpha}F(\balpha)=\nabla_{\!\balpha}\Lval(\bw^*\!,\balpha)+\nabla_{\!\balpha}^T\!\bw^*\!(\balpha)\nabla_{\!\bw}\Lval(\bw^*\!(\balpha),\balpha),
\end{equation}
where
\vspace{-3pt}
\begin{equation}\label{eq3}
    \nabla_{\!\balpha}\bw^*\!(\balpha)=-[\nabla_{\bw\bw}^2\Ltrain(\bw^{*},\balpha)]^{-1}\nabla_{\balpha\bw}^2\Ltrain(\bw^*,\balpha).
\end{equation}
Due to the impracticality of computing the Hessian as in Eq.~\eqref{eq3}, Xie et al. introduce ZO-DARTS~\cite{xie2023zo} to circumvent this obstacle. ZO-DARTS employs a zeroth-order approximation technique~\cite{nesterov2017random}:
\vspace{-3pt}
\begin{equation}\label{eq4}
\tilde{\nabla}_{\balpha}F(\balpha)\triangleq\frac{F(\balpha+\mu\bu)-F(\balpha)}{\mu}\bu.
\end{equation}
As $\mu\rightarrow0$, this approximates the directional derivative $\nabla_{\balpha}^T F(\balpha)\bu$. Combining Eq.~\eqref{eq2} and Eq.~\eqref{eq4} we have:
\vspace{-3pt}
\begin{equation}\label{eq5}
\scalebox{0.93}{$
\begin{aligned}
    \tilde{\nabla}_{\balpha}F(\balpha)
    \approx\nabla_{\balpha}^T F(\balpha)\bu\bu
    =&\nabla_{\!\balpha}^T\Lval(\bw^*\!,\!\balpha)\bu\bu+\nabla_{\!\bw}^T\Lval(\bw^*\!(\balpha),\!\balpha)\nabla_{\!\balpha}\bw^*\!(\balpha)\bu\bu\\
    =&\nabla_{\!\balpha}^T\Lval(\bw^*\!,\!\balpha)\bu\bu+[\nabla_{\!\balpha}\bw^*\!(\balpha)\bu]^T\nabla_{\!\bw}\Lval(\bw^*\!(\balpha),\balpha)\bu.
\end{aligned}
$}
\end{equation}
Thus, $\nabla_{\balpha}\bw^*(\balpha)\bu$ can be effectively approximated and computed swiftly using this method.

\section{The proposed method}\label{sec3}
\vspace{-5pt}

ZO-DARTS+ is an advanced NAS algorithm that improves efficiency by integrating a novel probability normalization function with an annealing strategy. Traditionally, softmax is used to generate probability parameters from a set of architecture variables, which are then used to compute a weighted sum of operations in DARTS. However, this method struggles with the complexity of bi-level NAS problems, as softmax tends to produce probabilities that do not sufficiently diverge and remain too similar across different operations. This similarity is detrimental to the selection of discrete operations, where a sparse probability distribution is preferable. To overcome these limitations, ZO-DARTS+ employs the sparsemax function~\cite{martins2016softmax}, which projects the input vector $\balpha$ onto the probability simplex to produce a sparser probability distribution:
\vspace{-5pt}
\begin{equation}\label{eq6}
    {\rm sparsemax}(\balpha):={\rm argmin}_{\bp\in\Delta^{K-1}}||\bp-\balpha||^2.
\end{equation}
The $(K-1)$-dimensional simplex, $\Delta^{K-1}:=\{\bp\in\mathbb{R}^K|\boldsymbol{1}^T\bp=1, \bp\geq0\}$, ensures that the sum of the probabilities is equal to one and each probability is non-negative, increasing the likelihood of obtaining a sparse result. This sparse distribution of the probabilities is used in the mixed operation as follows:
\vspace{-5pt}
\begin{equation}\label{eq7}
    \bar{o}^{(i,j)}(x)=\sum_{o\in\mathcal{O}}{\rm sparsemax}_o(\balpha)o(x).
\end{equation}
Sparsemax is designed to push architecture parameters to extremes (close to 1 for selected operations and 0 for others), but it does not always eliminate non-zero elements completely. To refine this, a temperature parameter $\tau$ is used to modulate the input values, thereby increasing sparsity:
\vspace{-5pt}
\begin{equation}\label{eq8}
    \mathrm{softmax}_o(\boldsymbol{\balpha}) = \frac{\exp(\balpha_o^{(i,j)}/\tau)}{\sum_{o'\in\mathcal{O}}\exp(\balpha_{o'}^{(i,j)}/\tau)}.
\end{equation}
As $\tau\rightarrow0$ this approach closely resembles an argmax function, ideal for achieving discrete selections. To further enhance this property, we propose an annealing strategy to reduce $\tau$ by a manually chosen factor $a \leq 1$ every $m$ epochs, driving the system towards a one-hot vector:
\vspace{-5pt}
\begin{equation}\label{eq9}
    \mathrm{sparsemax}(\balpha) = {\arg\max}_{\bp\in\Delta^{k-1}}||\bp - \balpha/(\tau * a^{n//m})||^2,
\end{equation}
 where $n$ is the total number of epochs, and $//$ indicates an integer division. By adjusting the factors $a$ and $m$, the algorithm efficiently accelerates the search process while promoting sparsity, closely matching the need for fast and automatic model generation.

\vspace{-10pt}
\section{Experiments and Results}\label{sec4}
\vspace{-5pt}

To demonstrate the effectiveness and efficiency of our approach, we conducted experiments on five medical image datasets from MedMNIST~\cite{medmnistv2}, namely PneumoniaMNIST, OCTMNIST, BreastMNIST, OrganAMNIST, and OrganSMNIST. In our NAS framework, both ZO-DARTS and ZO-DARTS+ update the architecture parameters $\balpha$ only when the model parameters $\bw$ are changed, which happens every $T=10$ rounds. An initial temperature $\tau=1.5$ is used to encourage exploration in the early stages of the search. The annealing factor is set to $a=0.75$, with an interval of $m=5$, allowing for a gradual refinement of the model configurations. For benchmarking, we used ResNet18 as a standard for manually designed models and included AutoKeras and Google AutoML Vision as baselines for NAS models. In addition, we evaluated three DARTS-style methods -- DARTS~\cite{liu2018darts}, MileNAS~\cite{he2020milenas}, and ZO-DARTS~\cite{xie2023zo} -- all of which operate within a search space defined by NAS-Bench-201~\cite{dong2020bench} with five operations: Zeroise, Skip Connect, 1x1 Conv, 3x3 Conv, and 3x3 Avg Pooling. We performed 50 search rounds for each method. The final model accuracy is determined by retraining to full convergence after each search. We conducted these experiments three times to get the average and standard deviation of performance, and the average search time.

\begin{table}[t]
    \centering
    \setlength{\tabcolsep}{2.5pt}
    \footnotesize
    \caption{Accuracy comparison with baselines and DARTS-style methods.}
    \label{tab:acc}
    \scalebox{0.88}{
    \begin{tabular}{lccccc}
        \toprule
        Dataset & PneumoniaMNIST &  OCTMNIST &  BreastMNIST &  OrganAMNIST &  OrganSMNIST \\ 
        \midrule
        ResNet18            & 0.854 &  0.743 &  0.863 &  \underline{0.935} &  0.782 \\ 
        AutoKeras           & 0.878 &  0.763 &  0.831 &  0.905 &  \textbf{0.813} \\ 
        Google AutoML       & \textbf{0.946} &  \underline{0.771} &  0.861 &  0.886 &  0.749 \\ 
        \midrule
        DARTS               &  0.935$\pm$0.008 &  0.756$\pm$0.028 &  0.876$\pm$0.010 &  \underline{0.935}$\pm$0.014 &  0.771$\pm$0.015 \\
        MiLeNAS               &  0.625$\pm$0.000 &  0.610$\pm$0.312 &  \textbf{0.885}$\pm$0.000 &  0.695$\pm$0.442 &  0.762$\pm$0.008 \\
        ZO-DARTS            &  \underline{0.941}$\pm$0.003 &  \textbf{0.807}$\pm$0.005 &  \underline{0.878}$\pm$0.013 &  0.928$\pm$0.011 &  0.787$\pm$0.004 \\ 
        ZO-DARTS+ (ours)    &  0.937$\pm$0.011 &  0.745$\pm$0.086 &  0.872$\pm$0.011 &  \textbf{0.939}$\pm$0.008 &  \underline{0.790}$\pm$0.013 \\ 
        \bottomrule
    \end{tabular}
    }
    \vspace{-15pt}
\end{table}

\begin{table}[t]
    \centering
    \setlength{\tabcolsep}{2.5pt}
    \footnotesize
    \caption{Search Time (second) comparison with DARTS-style methods.}
    \label{tab:time}
    \scalebox{0.88}{
    \begin{tabular}{lccccc}
        \toprule
        Dataset & PneumoniaMNIST &  OCTMNIST &  BreastMNIST &  OrganAMNIST &  OrganSMNIST \\ 
        \midrule
        % ResNet18 &  / &  / &  / &  / &  / \\ 
        % AutoKeras &  - &  - &  - &  - &  - \\ 
        % Google AutoML & - &  - &  - &  - &  - \\ 
        DARTS               & 1990.3 &  4217.1 &  503.9 &  2173.8 &  754.2 \\
        MiLeNAS               &  1726.4 &  3436.1 &  390.7 &  1895.1 &  644.0 \\
        ZO-DARTS            &  825.5 &  1792.3 &  213.7 &  874.8 &  354.6 \\ 
        ZO-DARTS+ (ours)    & \textbf{688.0} &  \textbf{1523.8} &  \textbf{170.1} &  \textbf{717.0} &  \textbf{298.9} \\ 
        \bottomrule
    \end{tabular}
    }
    \vspace{-15pt}
\end{table}

Table~\ref{tab:acc} shows the optimal performance achieved by each algorithm on the different datasets, with baseline data taken directly from the MedMNIST homepage\footnote{https://medmnist.com}. ZO-DARTS+ outperforms ResNet-18 on five datasets and both Google AutoML and AutoKeras on three. It also consistently ranks first/second or close to the best among DARTS-style methods, demonstrating robust performance across the board. Despite this narrow accuracy gap between ZO-DARTS+ and other DARTS-style methods, there is a significant reduction in search time. Surprisingly, ZO-DARTS+ tends to converge and stop the search by early stopping around the $40^{th}$ epoch of search, further reducing the search time. As detailed in Table~\ref{tab:time}, this efficiency underlines the suitability of ZO-DARTS+ for the development of medical image classification models, balancing state-of-the-art accuracy with up to a threefold reduction in search time.

\begin{figure}[t]
    \centering
    \includegraphics[width=1.\textwidth]{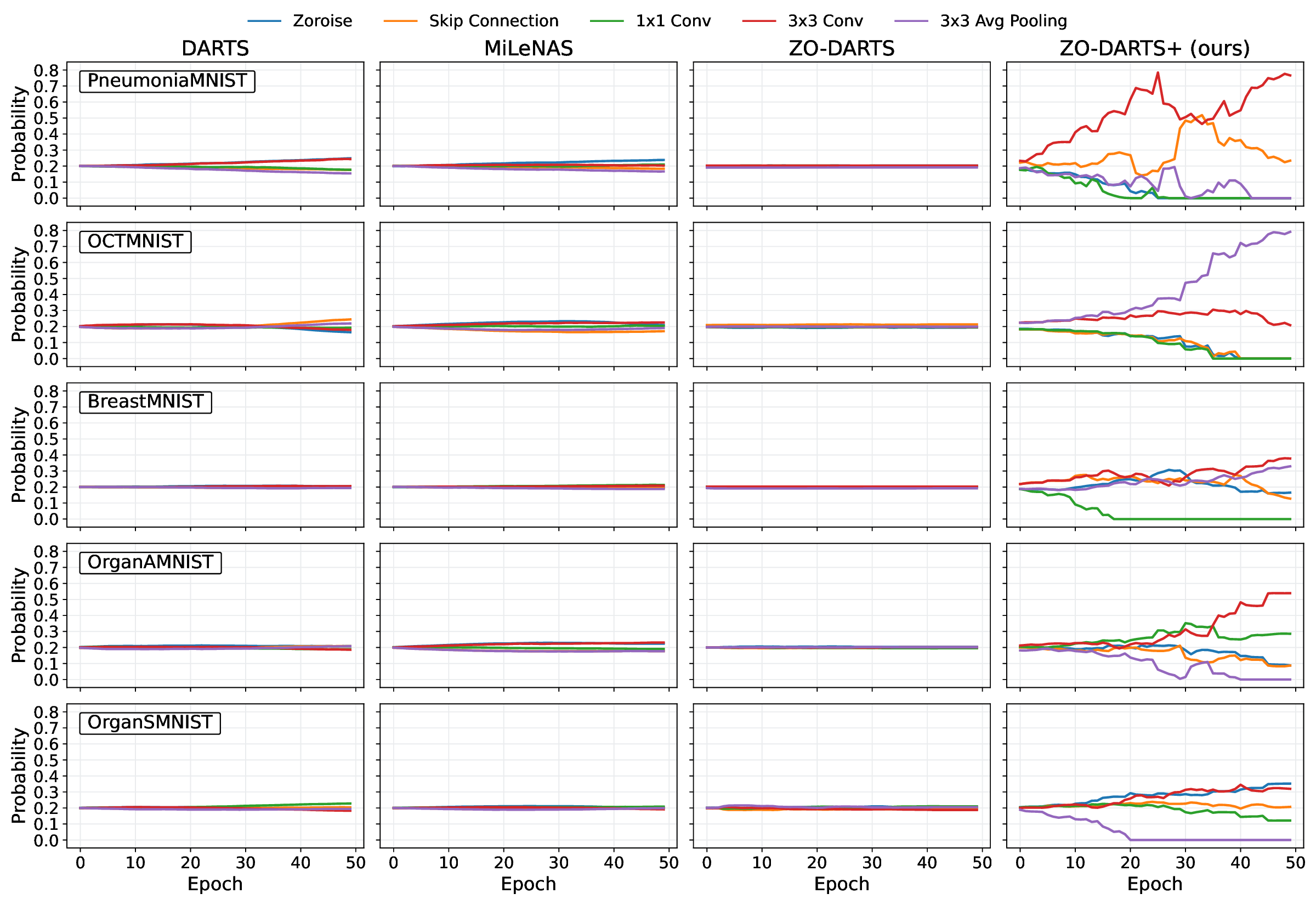}
    \vspace{-20pt}
    \caption{Probability rank variation of one edge during the search procedure.}
    \label{fig:1}
    \vspace{-10pt}
\end{figure}

We analyzed the progression of architecture operator parameters across the different NAS methods and plotted these progressions in Fig~\ref{fig:1}. All models started optimization with the same initial probability weights. Throughout the search process, most NAS methods exhibited limited variability in these weights, leading to challenges in reliably selecting the most effective operation -- the probabilities remained too close together, allowing for easy rank changes, thus compromising the interpretability of the operators.
Unlike other methods, ZO-DARTS+ showed rapid convergence in selecting the best operation, thanks to the integration of sparsemax and an annealing strategy. This efficiency not only improves performance, but also avoids the high costs associated with increasing the number of search rounds to achieve clear probability separations.

In addition, the analysis revealed distinct preferences for certain operations of one edge depending on the dataset. For example, the 3x3 Conv operation was frequently selected for the PneumoniaMNIST dataset, whereas the 3x3 Avg Pooling was predominant in the OCTMNIST dataset, with other operations often excluded from the final selection altogether. This pattern highlights the adaptability of ZO-DARTS+, allowing it to effectively tailor its architecture to different types of data. These findings -- coupled with robust accuracy results and reduced search times -- confirm that ZO-DARTS+ is a highly validated model that demonstrates significant adaptability and reliability in its operation selection, making it a superior choice for medical image classification tasks.

\vspace{-10pt}

\section{Conclusions and Future works}\label{sec5}
\vspace{-5pt}

This paper presents ZO-DARTS+, a novel differentiable NAS algorithm that achieves up to three times faster search times compared to existing leading methods, while preserving accuracy on medical image datasets. These results validate our hypothesis and are achieved through the integration of the sparsemax function and a tailored annealing strategy during operator selection, resulting in sparser and more efficient solutions. Future research will build on this foundation by applying stringent constraints to reduce both computational resources and time further. In addition, the application of sparsity-aware annealing strategies in different NAS frameworks will be explored to validate the effectiveness of the approach in different settings. 

\vspace{-10pt}
\bibliographystyle{unsrt}
\bibliography{reference}
\end{document}